\documentclass[10pt,twocolumn]{article}

\usepackage[margin=0.75in,top=0.55in]{geometry}
\usepackage{etoolbox}
\usepackage{amsmath,amssymb,amsfonts}
\usepackage{graphicx}
\usepackage{booktabs}
\usepackage{hyperref}
\usepackage{algorithm}
\usepackage{algpseudocode}
\usepackage{tikz}
\usetikzlibrary{shapes.geometric, arrows.meta, positioning, fit, calc, backgrounds}
\usepackage{xcolor}
\usepackage{caption}
\usepackage{subcaption}
\usepackage{enumitem}
\usepackage{microtype}
\usepackage{authblk}
\usepackage{fancyhdr}
\usepackage{mathtools}
\usepackage{listings}
\lstdefinestyle{archascii}{
  basicstyle=\ttfamily\footnotesize,
  frame=single,
  rulecolor=\color{gray!50},
  columns=fullflexible,
  keepspaces=true,
  aboveskip=5pt,
  belowskip=5pt,
  lineskip=-1pt
}
\usepackage{titlesec}
\titleformat{\section}
  {\normalfont\bfseries\fontsize{11}{13}\selectfont\raggedright}
  {\thesection}{0.55em}{}
\titlespacing*{\section}{0pt}{1.1ex plus .15ex}{0.65ex}
\titleformat{\subsection}
  {\normalfont\bfseries\normalsize\raggedright}
  {\thesubsection}{0.45em}{}
\titlespacing*{\subsection}{0pt}{0.85ex plus .12ex}{0.45ex}
\titleformat{\subsubsection}
  {\normalfont\bfseries\small\raggedright}
  {\thesubsubsection}{0.45em}{}
\titlespacing*{\subsubsection}{0pt}{0.7ex plus .1ex}{0.35ex}

\definecolor{darkblue}{RGB}{0,51,102}
\definecolor{lightgray}{RGB}{245,245,245}
\definecolor{accentblue}{RGB}{52,119,198}
\definecolor{accentgreen}{RGB}{46,139,87}
\definecolor{accentorange}{RGB}{210,105,30}

\hypersetup{
    colorlinks=true,
    linkcolor=darkblue,
    citecolor=darkblue,
    urlcolor=accentblue
}

\title{\textbf{Object-Centric Stereo Ranging for Autonomous Driving:\\
From Dense Disparity to Census-Based Template Matching}}

\author{Qihao Huang, With Cursor}
\date{}

\makeatletter
\patchcmd{\@maketitle}{\vskip 2em}{\vskip 0.45em}{}{}
\patchcmd{\@maketitle}{\vskip 1.5em}{\vskip 0.45em}{}{}
\patchcmd{\@maketitle}{\vskip 1.5em}{\vskip 0.45em}{}{}
\patchcmd{\@maketitle}{\vskip 1em}{\vskip 0.3em}{}{}
\makeatother

\begin{document}
\maketitle

\begin{abstract}
Accurate depth estimation is critical for autonomous driving perception systems, particularly for long-range vehicle detection on highways.
Traditional dense stereo matching methods such as Block Matching (BM) and Semi-Global Matching (SGM) produce per-pixel disparity maps but suffer from computational cost, sensitivity to radiometric differences between stereo cameras, and poor accuracy at long ranges where disparity values are small.
In this report, we present a comprehensive stereo ranging system that integrates three complementary depth estimation approaches---dense BM/SGM disparity, object-centric Census-based Template Matching, and monocular geometric priors---within a unified detection-ranging-tracking pipeline.
Our key contribution is a novel \emph{object-centric Census-based Template Matching} algorithm that performs GPU-accelerated sparse stereo matching directly within detected bounding boxes, employing a far/close divide-and-conquer strategy, forward-backward verification, occlusion-aware sampling, and robust multi-block aggregation.
We further describe an online calibration refinement framework that combines auto-rectification offset search, radar-stereo voting-based disparity correction, and object-level radar-stereo association for continuous extrinsic drift compensation.
The complete system achieves real-time performance through asynchronous GPU pipeline design and delivers robust ranging across diverse driving conditions including nighttime, rain, and varying illumination.
\end{abstract}

\section{Introduction}
\label{sec:intro}

Reliable 3D perception is the foundation of autonomous driving.
Among the available sensing modalities, stereo vision occupies a unique position: it provides \emph{physics-grounded} metric depth through triangulation---unlike monocular vision which requires learned priors or heuristic assumptions---while using only passive cameras at a fraction of the cost of LiDAR.
This combination has led stereo cameras to be described as a ``pseudo-LiDAR''~\cite{wang2019pseudo}: a system that, like LiDAR, produces dense 3D point clouds, but relies solely on off-the-shelf imaging sensors.

Given a calibrated stereo camera pair with baseline $b$ and focal length $f$, the depth $Z$ of a point is determined by its disparity $d$:
\begin{equation}
\label{eq:depth_from_disparity}
Z = \frac{f \cdot b}{d}
\end{equation}
This inverse relationship implies that depth accuracy degrades quadratically with distance, making long-range estimation particularly challenging for highway scenarios where vehicles must be detected and ranged beyond 200 meters.
Yet it is precisely in this regime---long-range, high-speed highway driving---that stereo vision proves indispensable, because it delivers direct geometric depth without requiring any prior knowledge about the objects being observed.

\subsection{The Role of Stereo Vision in AD}

To appreciate why stereo vision matters, it is instructive to compare the dominant depth sensing paradigms in autonomous driving:

\paragraph{Monocular Vision.}
Single-camera depth estimation is fundamentally ill-posed: recovering 3D structure from a 2D projection requires either learned priors (deep networks trained on large datasets) or geometric assumptions (known object sizes, ground plane constraints).
Monocular methods follow a ``detect first, then infer depth'' pipeline---they must \emph{recognize} an object before estimating its distance.
This creates a critical blind spot for \textbf{non-standard obstacles} (fallen cargo, road debris, unusual vehicles) that were never seen during training.
Typical monocular depth errors range from 15--30\%, and the error increases significantly beyond 50m where perspective cues become weak~\cite{fu2018deep}.

\paragraph{LiDAR.}
Active laser scanning provides centimeter-level range accuracy and is independent of ambient illumination.
However, LiDAR suffers from (a) high cost (\$450--\$8000 per unit), (b) sparse point returns at long range (the angular resolution of a 128-line LiDAR yields only a few points on a vehicle at 200m), (c) degradation in adverse weather---recent comparative studies~\cite{nodar2025} report that LiDAR valid data rates drop to 40\% in heavy rain and 20\% in fog, while stereo vision maintains 70\% in both conditions, and (d) vulnerability to mutual interference in high-traffic scenarios.

\paragraph{Stereo Vision.}
Binocular stereo computes depth from pixel-level correspondences between two synchronized views.
Its key advantages for autonomous driving are:
\begin{itemize}[leftmargin=*,nosep]
\item \textbf{Physics-based depth}: Depth is derived from the triangulation equation (Eq.~\ref{eq:depth_from_disparity}), requiring no learned priors and no object recognition. This enables ranging of \emph{arbitrary} objects---including non-standard obstacles that monocular systems cannot handle.
\item \textbf{Dense 3D}: A stereo pair can produce up to $50$ million depth measurements per second (10--50$\times$ more than high-end LiDAR), enabling detection of small objects at extreme range.
\item \textbf{Low cost}: Camera modules cost \$30--80, roughly 10--100$\times$ cheaper than LiDAR, making stereo vision viable for mass-market vehicles.
\item \textbf{Weather robustness}: The Census Transform and other non-parametric matching costs are inherently invariant to gain and exposure differences, providing robustness to rain, fog, low-light, and inter-camera radiometric variation.
\item \textbf{Hardware synchronization}: Stereo pairs provide a natural reference clock for the perception pipeline. In a vision-centric architecture, the stereo trigger can synchronize all other cameras, yielding temporally consistent multi-view data.
\end{itemize}

\subsection{Long-Range Perception}

For highway trucking at 100 km/h, a safe following distance of 3 seconds corresponds to $\sim$83m, but anticipatory planning requires detecting and tracking vehicles at 200--300m or beyond.
At these ranges, stereo vision faces its fundamental challenge: the expected disparity shrinks to the sub-pixel regime.
For a typical truck-mounted stereo camera ($b = 30$cm, $f = 2000$px), a target at 200m produces a disparity of only $d = fb/Z = 3.0$ pixels.
Dense matching methods (BM, SGM), which operate on integer or coarse sub-pixel grids, struggle to produce reliable estimates at this scale.

Our key insight is that \textbf{object-centric sparse matching can achieve superior long-range accuracy} compared to dense methods, because:
\begin{enumerate}[leftmargin=*,nosep]
\item Concentrating all query points within a detected bounding box aggregates evidence from hundreds of pixels, effectively ``voting'' for the object's disparity with far greater signal-to-noise ratio than any single pixel.
\item Census-based matching is invariant to monotonic intensity changes, eliminating the radiometric calibration errors that plague SAD/SSD-based dense methods.
\item Forward-backward verification rejects unreliable matches at the algorithmic level, rather than relying on post-hoc filtering of a noisy disparity map.
\item Sub-pixel parabolic interpolation on the aggregated cost recovers fractional disparity with precision well below 1 pixel.
\end{enumerate}

\subsection{Comparison with Alternative Ranging Approaches}

Table~\ref{tab:sensing_comparison} positions stereo vision among the major depth sensing modalities used in autonomous driving.

\begin{table}[t]
\centering
\caption{Comparison of depth sensing modalities for autonomous driving. Stereo vision uniquely combines physics-based metric depth, low cost, and dense output.}
\label{tab:sensing_comparison}
\footnotesize
\begin{tabular}{@{}lcccc@{}}
\toprule
& \textbf{Mono} & \textbf{Stereo} & \textbf{LiDAR} & \textbf{Radar} \\
\midrule
Depth type & Learned & Geometric & Active & Active \\
Cost (\$) & 10--30 & 30--80 & 450--8k & 50--200 \\
Range (m) & $<$80 & $<$300+ & $<$200 & $<$250 \\
Density & Dense & Dense & Sparse & Very sparse \\
Non-std obj. & Poor & Good & Good & Poor \\
Weather & Moderate & Good & Poor & Excellent \\
Metric acc. & Low & High & Very high & Moderate \\
\bottomrule
\end{tabular}
\end{table}

The critical distinction is that \textbf{stereo and LiDAR are the only modalities that provide physics-based metric depth without requiring object recognition}, making them essential for safety-critical perception.
Where LiDAR excels in absolute accuracy, stereo excels in density, cost, and weather robustness---making the two highly complementary.
Monocular depth, while computationally cheap, relies on data-driven priors that may fail silently on out-of-distribution objects.
Radar provides excellent velocity and weather resilience but lacks the spatial resolution for precise 3D localization.

\subsection{Contributions and Outline}

Classical approaches to stereo matching---Block Matching (BM)~\cite{konolige1998small} and Semi-Global Matching (SGM)~\cite{hirschmuller2005accurate}---compute dense disparity maps across the entire image.
While effective for close-range obstacles, these methods face several limitations in production autonomous driving systems:

\begin{enumerate}[leftmargin=*,nosep]
\item \textbf{Computational cost}: Dense matching over high-resolution images ($1920 \times 1200$) is expensive even with GPU acceleration, competing for resources with neural network inference.
\item \textbf{Radiometric sensitivity}: Left-right camera pairs may exhibit different gain, exposure, or color response, degrading correlation-based matching.
\item \textbf{Long-range accuracy}: For example, at 200m with a 12cm baseline and 2000px focal length, the expected disparity is merely $\sim$1.2 pixels, well below the noise floor of dense methods (cf.\ the 30cm-baseline example in Section~\ref{sec:intro}).
\item \textbf{Wasted computation}: For object-level ranging in autonomous driving, only the disparity at detected object locations is needed, not a full disparity map.
\end{enumerate}

To address these limitations, we propose an \textbf{object-centric Census-based Template Matching} approach that fundamentally changes the matching paradigm: instead of computing a dense disparity map, we perform sparse stereo matching \emph{only at query points sampled within detected bounding boxes}.
By combining the Census Transform's robustness to radiometric variations with GPU-parallel block matching and multi-scale processing, our method achieves both superior accuracy at long range and lower computational cost.

This report presents the complete system architecture, from detection to ranging, online calibration refinement, and temporal fusion through vision tracking.
Section~\ref{sec:related} reviews related work.
Section~\ref{sec:architecture} describes the system architecture.
Section~\ref{sec:methods} details the three stereo matching methods.
Section~\ref{sec:calibration} presents our online calibration refinement framework.
Section~\ref{sec:tracking} describes the vision tracking and fusion pipeline.
Section~\ref{sec:discussion} provides system design discussion, and Section~\ref{sec:conclusion} concludes.

\section{Related Work}
\label{sec:related}

\paragraph{Dense Stereo Matching.}
Block Matching (BM)~\cite{konolige1998small} computes disparity by sliding a fixed-size window across the epipolar line, using Sum of Absolute Differences (SAD) as the cost metric.
Semi-Global Matching (SGM)~\cite{hirschmuller2005accurate} extends pixel-wise matching with smoothness constraints aggregated along multiple scanline directions, producing significantly smoother disparity maps.
GPU-accelerated implementations~\cite{hernandez2016embedded} have made both methods viable for real-time applications, though at significant computational cost.

\paragraph{Census Transform.}
The Census Transform~\cite{zabih1994non} encodes local structure by comparing each pixel's intensity with its neighbors, producing a binary string that is invariant to monotonic intensity changes.
Hamming distance between Census descriptors provides a matching cost that is robust to radiometric differences between cameras.
The AD-Census method~\cite{mei2011building} combines Census with absolute differences for improved accuracy.

\paragraph{Object-Level and Detection-Driven Stereo.}
Rather than computing dense disparity, several works have explored object-level stereo matching for autonomous driving.
3DOP~\cite{chen20153d} generates 3D proposals from stereo depth maps, while Stereo R-CNN~\cite{li2019stereo} jointly detects objects in left-right images and estimates 3D bounding boxes.
The Pseudo-LiDAR framework~\cite{wang2019pseudo} converts stereo depth maps into 3D point clouds and applies LiDAR-based detectors, demonstrating that the representation format---not the depth source---is the primary bottleneck.
Our work takes a different approach: rather than converting dense disparity to point clouds, we perform \emph{sparse object-level matching} directly, avoiding the computational overhead of dense stereo entirely.

\paragraph{Monocular 3D Perception.}
Monocular 3D detection methods~\cite{fu2018deep,brazil2019m3d} estimate depth from single images using learned priors, geometric constraints (ground plane, object size), or direct regression.
While computationally efficient, these methods are inherently limited by the ill-posed nature of monocular depth and exhibit scale ambiguity, particularly for novel object categories.
BEV-based methods~\cite{philion2020lift,li2022bevformer} lift 2D features to 3D using predicted depth distributions, but still rely on learned depth priors that may not generalize to rare or non-standard obstacles.

\paragraph{Sensor Fusion for Calibration.}
Online stereo calibration refinement has been addressed through feature matching~\cite{dang2009continuous} and radar-camera fusion~\cite{wang2021radar}.
Our system combines multiple complementary refinement signals---disparity-maximizing rectification search, radar-based voting, and object-level radar-stereo association---for robust continuous calibration.

\section{System Architecture}
\label{sec:architecture}

\begin{figure*}[t]
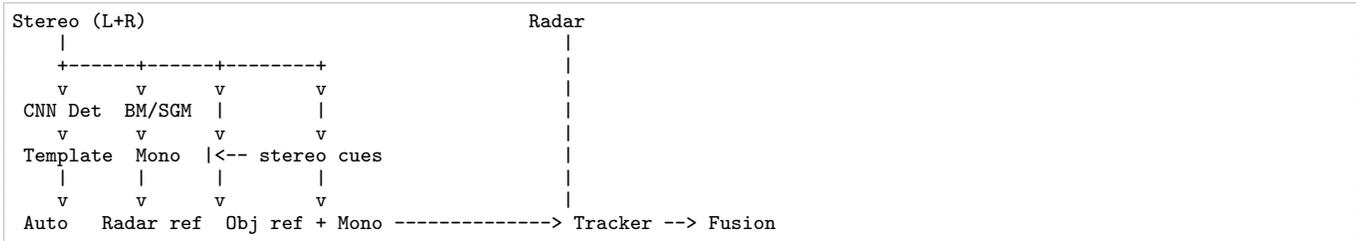

\centering
\begin{minipage}{\textwidth}
\begin{lstlisting}[style=archascii]
Stereo (L+R)                                  Radar
    |                                            |
    +------+------+--------+                     |
    v      v      v        v                     |
 CNN Det  BM/SGM  |        |                     |
    v      v      v        v                     |
 Template  Mono  |<-- stereo cues                |
    |      |      |        |                     |
    v      v      v        v                     |
 Auto   Radar ref  Obj ref + Mono --------------> Tracker --> Fusion
\end{lstlisting}
\end{minipage}
\caption{ASCII schematic of the stereo ranging pipeline (read top to bottom). Parallel branches: CNN$\to$Template Match and BM/SGM$\to$Mono; stereo also feeds monocular cues. TM and BM/SGM feed Auto Rectification; TM also feeds Object Disparity Refiner; radar and mono feed both refiners; mono enters the tracker. Refiners precede Hungarian-matched Kalman tracking and fusion.}
\label{fig:architecture}
\end{figure*}

The system processes stereo image pairs through a multi-stage pipeline with extensive parallelism, as illustrated in Figure~\ref{fig:architecture}. The design philosophy is \emph{compute depth through multiple independent methods, then fuse results with appropriate uncertainty estimates}.

\subsection{Pipeline Overview}

Upon receiving a synchronized stereo image pair, the system launches several concurrent GPU operations:

\begin{enumerate}[leftmargin=*,nosep]
\item \textbf{Object Detection}: A CNN-based detector runs on the left image, producing 2D bounding boxes with class labels and keypoints.
\item \textbf{Dense Disparity} (if configured): BM or SGM computes a full disparity map from rectified grayscale images.
\item \textbf{Template Matching} (if configured): Object-level Census-based matching computes per-detection disparity.
\item \textbf{Monocular Depth}: Ground plane projection and apparent size provide complementary depth estimates.
\end{enumerate}

These operations overlap through CUDA stream management. Detection and disparity computation proceed in parallel on separate GPU streams, with synchronization only at the point of fusion.

\subsection{Depth Method Selection}

The system supports three primary depth methods, selected at initialization:
\begin{itemize}[leftmargin=*,nosep]
\item \texttt{STEREO\_BM}: Block Matching dense disparity
\item \texttt{STEREO\_SGM}: Semi-Global Matching dense disparity
\item \texttt{TEMPLATE\_MATCHER}: Object-centric Census matching
\end{itemize}

Regardless of the primary stereo method, monocular depth cues are always computed as fallback and for cross-validation.
The closest point result structure maintains separate estimates: \texttt{clp\_by\_stereo} (from the primary stereo method), \texttt{clp\_by\_gpt} (ground point triangulation), and \texttt{clp\_by\_size} (apparent size regression).
A priority-based selection with sanity checking determines the final depth estimate for each detection.

\section{Stereo Matching Methods}
\label{sec:methods}

\begin{figure*}[t]
\centering
\begin{tikzpicture}[
    scale=0.88, every node/.style={scale=0.88},
    imgframe/.style={rectangle, draw=gray!50, thick, minimum width=3.0cm, minimum height=2.0cm, fill=gray!6},
    outbox/.style={rectangle, rounded corners=3pt, draw, thick, minimum width=2.6cm, minimum height=0.5cm, font=\scriptsize, align=center},
    arr/.style={-{Stealth[length=3pt]}, thick, gray!60},
    paneltitle/.style={font=\small\bfseries},
]

\begin{scope}[shift={(0,0)}]
\node[paneltitle, text=darkblue] at (0, 1.6) {(a) Block Matching};
\node[imgframe] (bm) at (0, 0) {};
\node[font=\tiny, gray] at (0, -1.25) {Left Image};
\fill[accentblue!10] (-1.35, -0.85) rectangle (1.35, 0.85);
\foreach \y in {0.55, 0.25, -0.05, -0.35, -0.65} {
    \draw[accentblue!30, thin, dashed] (-1.3, \y) -- (1.3, \y);
}
\draw[accentorange, very thick, fill=accentorange!25] (-0.35, -0.1) rectangle (0.45, 0.5);
\node[font=\tiny\bfseries, accentorange!80!black] at (0.05, 0.2) {SAD};
\draw[accentorange, thick, -{Stealth[length=3pt]}] (0.5, 0.2) -- (1.25, 0.2);
\node[font=\tiny, accentorange!70!black, rotate=0] at (0.87, 0.38) {search};
\draw[arr] (0, -1.55) -- (0, -2.15);
\node[outbox, fill=accentblue!8, draw=accentblue!50] at (0, -2.55) {Dense Disparity Map};
\node[font=\tiny, gray!70, align=center] at (0, -3.15) {$\forall$ pixels, $O(W{\times}H{\times}N_d)$};
\end{scope}

\begin{scope}[shift={(5.2,0)}]
\node[paneltitle, text=darkblue] at (0, 1.6) {(b) Semi-Global Matching};
\node[imgframe] (sgm) at (0, 0) {};
\node[font=\tiny, gray] at (0, -1.25) {Left Image};
\fill[accentgreen!8] (-1.35, -0.85) rectangle (1.35, 0.85);
\coordinate (cp) at (0, 0);
\foreach \a/\lbl in {0/\rightarrow, 45/, 90/\uparrow, 135/, 180/\leftarrow, 225/, 270/\downarrow, 315/} {
    \draw[accentgreen!70, thick, -{Stealth[length=2.5pt]}]
        ([shift=(\a:0.75cm)]cp) -- ([shift=(\a:0.12cm)]cp);
}
\fill[accentgreen!80!black] (cp) circle (2.5pt);
\node[font=\tiny, accentgreen!60!black] at (0.55, 0.55) {$P_1,P_2$};
\draw[arr] (0, -1.55) -- (0, -2.15);
\node[outbox, fill=accentgreen!8, draw=accentgreen!50!black] at (0, -2.55) {Dense Disparity Map};
\node[font=\tiny, gray!70, align=center] at (0, -3.15) {$\forall$ pixels, $O(W{\times}H{\times}N_d{\times}K)$};
\end{scope}

\begin{scope}[shift={(10.4,0)}]
\node[paneltitle, text=accentorange!85!black] at (0.2, 1.6) {(c) Template Match \textbf{(Ours)}};
\node[imgframe] (tm) at (0, 0) {};
\node[font=\tiny, gray] at (0, -1.25) {Left Image + Detections};
\draw[red!70, very thick, fill=red!12] (-1.1, 0.25) rectangle (-0.2, 0.75);
\node[font=\tiny\bfseries, red!60!black] at (-0.65, 0.50) {Far};
\draw[red!70, very thick, fill=red!12] (0.1, -0.65) rectangle (1.2, 0.15);
\node[font=\tiny\bfseries, red!60!black] at (0.65, -0.25) {Close};
\draw[red!70, thick, fill=red!12] (-0.7, -0.6) rectangle (-0.15, -0.2);
\node[font=\tiny\bfseries, red!60!black] at (-0.42, -0.4) {\scriptsize F};
\node[font=\tiny, blue!60!black] at (-0.65, 0.32) {\texttt{01101}};
\node[font=\tiny, blue!60!black] at (0.65, -0.50) {\texttt{10011}};
\draw[arr] (0, -1.55) -- (0, -2.15);
\node[outbox, fill=accentorange!12, draw=accentorange!70, very thick] at (0, -2.55) {Per-Object Disparity};
\node[font=\tiny, gray!70, align=center] at (0, -3.15) {Boxes only, $O(N_{\text{obj}}{\times}n{\times}N_d)$};
\draw[accentorange!70, very thick, rounded corners=6pt, densely dashed]
    (-1.7, -3.55) rectangle (2.1, 2.0);
\end{scope}

\begin{scope}[shift={(0,-4.4)}]
\node[font=\scriptsize\bfseries, gray!80] at (-0.5, 0) {Pixels processed:};
\fill[accentblue!30] (1.5, -0.15) rectangle (4.5, 0.15);
\node[font=\tiny, white] at (3.0, 0) {\textbf{100\% image}};

\fill[accentgreen!30] (6.7, -0.15) rectangle (9.7, 0.15);
\node[font=\tiny, white] at (8.2, 0) {\textbf{100\% image}};

\fill[accentorange!50] (11.2, -0.15) rectangle (12.2, 0.15);
\node[font=\tiny, white] at (11.7, 0) {\textbf{1--5\%}};
\end{scope}

\end{tikzpicture}
\caption{Fundamental paradigm comparison of three stereo matching methods. (a)~Block Matching slides a fixed SAD window along epipolar lines for \emph{every} pixel in the image. (b)~Semi-Global Matching aggregates pixel-wise matching costs along 4--8 directions with smoothness penalties ($P_1, P_2$) for \emph{every} pixel. Both produce dense disparity maps at $O(WH)$ cost. (c)~Our Template Matching (\textbf{highlighted}) performs Census-based matching \emph{only within detected bounding boxes}, processing 1--5\% of image pixels while achieving higher accuracy at long range through evidence aggregation.}
\label{fig:paradigm}
\end{figure*}
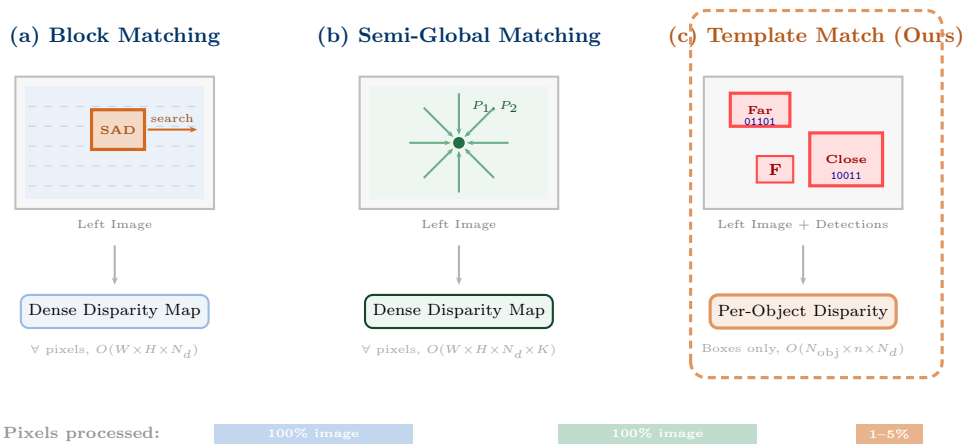

Figure~\ref{fig:paradigm} illustrates the fundamental paradigm difference among the three stereo methods.

\subsection{Block Matching (BM)}
\label{sec:bm}

The BM estimator wraps a CUDA-accelerated implementation with configurable parameters:
\begin{itemize}[leftmargin=*,nosep]
\item Number of disparities $N_d$ (search range)
\item Block size $w$ (matching window)
\item Minimum disparity $d_{\min}$
\item Texture threshold $\tau_t$ (reject low-texture regions)
\item Uniqueness ratio $\rho$ (reject ambiguous matches)
\end{itemize}

The input images are first converted to grayscale and optionally downsampled by a configurable factor $s$ to reduce computation.
When downsampling is applied, the resulting disparity map is upscaled by nearest-neighbor interpolation and the disparity values are multiplied by $s$ to recover the correct scale:
\begin{equation}
d_{\text{full}} = s \cdot d_{\text{down}}
\end{equation}

Invalid pixels (those below $d_{\min}$) are masked after upscaling to prevent artifacts.
The disparity map is stored in 16-bit fixed-point format (16 sub-pixel levels per pixel) for sub-pixel precision.

\subsection{Semi-Global Matching (SGM)}
\label{sec:sgm}

SGM extends pixel-wise matching with a smoothness prior enforced along multiple directions.
The energy function for a disparity assignment $D$ is:
\begin{multline}
E(D) = \sum_{\mathbf{p}} \Big( C(\mathbf{p}, D_\mathbf{p}) \\
+ \sum_{\mathbf{q} \in N_\mathbf{p}} P_1 \cdot \mathbb{1}[|D_\mathbf{p} - D_\mathbf{q}|{=}1] \\
+ P_2 \cdot \mathbb{1}[|D_\mathbf{p} - D_\mathbf{q}|{>}1] \Big)
\end{multline}
where $C(\mathbf{p}, d)$ is the pixel-wise matching cost, $P_1$ penalizes small disparity changes (slanted surfaces), and $P_2$ penalizes larger jumps (depth discontinuities).
The system uses the \texttt{MODE\_HH4} variant, which aggregates costs along 4 directions (horizontal, vertical, and two diagonals) for a balance between accuracy and speed.

\subsection{Census-Based Template Matching}
\label{sec:template_match}

This is our core contribution: an object-centric stereo matching method designed specifically for the autonomous driving ranging task.
Rather than computing a dense disparity map, we match \emph{only at points within detected object bounding boxes}, using the Census Transform for robustness to radiometric variations.

\subsubsection{Census Transform}

The Census Transform encodes local image structure into a binary descriptor.
For a pixel at position $(x, y)$ with intensity $I(x,y)$, the Census descriptor $\mathcal{C}(x,y)$ over a window of span $(S_x, S_y)$ is:
\begin{equation}
\mathcal{C}(x,y) = \bigoplus_{i=-S_x}^{S_x} \bigoplus_{j=-S_y}^{S_y} \xi\big(I(x+i, y+j), I(x,y)\big)
\end{equation}
where $\xi(a,b) = 1$ if $a > b$ and $0$ otherwise, and $\oplus$ denotes bitwise concatenation.
We use $S_x = S_y = 2$, producing a 25-bit descriptor (with 1 sentinel bit to distinguish from undefined regions, yielding 26 bits stored in a \texttt{uint32}).

\begin{figure}[t]
\centering
\begin{tikzpicture}[scale=0.78, every node/.style={scale=0.78}]
\node[font=\scriptsize\bfseries, darkblue] at (1.2, 3.1) {$5{\times}5$ Intensity Patch};
\def\vals{
    {48, 72, 35, 91, 63},
    {85, 57, 44, 68, 29},
    {61, 93, 55, 37, 76},
    {42, 66, 81, 50, 88},
    {73, 38, 59, 94, 46}
}
\foreach \r [count=\ri from 0] in {0,...,4} {
    \foreach \c [count=\ci from 0] in {0,...,4} {
        \pgfmathsetmacro{\val}{{\vals}[\r][\c]}
        \pgfmathtruncatemacro{\ival}{\val}
        \ifnum\ri=2
            \ifnum\ci=2
                \fill[accentorange!35] (\c*0.55, 2.2-\r*0.55) rectangle ++(0.5, 0.5);
                \draw[accentorange, thick] (\c*0.55, 2.2-\r*0.55) rectangle ++(0.5, 0.5);
            \else
                \pgfmathtruncatemacro{\cmp}{\ival > 55 ? 1 : 0}
                \ifnum\cmp=1
                    \fill[accentgreen!15] (\c*0.55, 2.2-\r*0.55) rectangle ++(0.5, 0.5);
                \else
                    \fill[red!10] (\c*0.55, 2.2-\r*0.55) rectangle ++(0.5, 0.5);
                \fi
                \draw[gray!40] (\c*0.55, 2.2-\r*0.55) rectangle ++(0.5, 0.5);
            \fi
        \else
            \pgfmathtruncatemacro{\cmp}{\ival > 55 ? 1 : 0}
            \ifnum\cmp=1
                \fill[accentgreen!15] (\c*0.55, 2.2-\r*0.55) rectangle ++(0.5, 0.5);
            \else
                \fill[red!10] (\c*0.55, 2.2-\r*0.55) rectangle ++(0.5, 0.5);
            \fi
            \draw[gray!40] (\c*0.55, 2.2-\r*0.55) rectangle ++(0.5, 0.5);
        \fi
        \node[font=\tiny] at (\c*0.55+0.25, 2.2-\r*0.55+0.25) {\ival};
    }
}
\draw[accentorange, thick, -{Stealth[length=3pt]}] (1.35, -0.7) -- (1.35, -0.3);
\node[font=\tiny\bfseries, accentorange!80!black] at (1.35, -0.9) {center=55};

\fill[accentgreen!15] (3.0, 2.15) rectangle ++(0.3, 0.3);
\draw[gray!40] (3.0, 2.15) rectangle ++(0.3, 0.3);
\node[font=\tiny, anchor=west, accentgreen!70!black] at (3.38, 2.30) {$>55 \to 1$};
\fill[red!10] (3.0, 1.6) rectangle ++(0.3, 0.3);
\draw[gray!40] (3.0, 1.6) rectangle ++(0.3, 0.3);
\node[font=\tiny, anchor=west, red!60!black] at (3.38, 1.75) {$\leq 55 \to 0$};

\draw[thick, gray!60, -{Stealth[length=3pt]}] (3.0, 0.6) -- (4.2, 0.6);
\node[font=\tiny, gray] at (3.6, 0.85) {encode};

\node[font=\scriptsize\bfseries, darkblue] at (5.8, 2.05) {Census Code};
\node[font=\footnotesize, blue!70!black, align=center] at (5.8, 0.45) {\texttt{0\,1\,0\,1\,1}\\[2pt]\texttt{1\,1\,0\,1\,0}\\[2pt]\texttt{1\,1\,\,\textbullet\,\,0\,1}\\[2pt]\texttt{0\,1\,1\,0\,1}\\[2pt]\texttt{1\,0\,1\,1\,0}};

\draw[thick, gray!60, -{Stealth[length=3pt]}] (5.8, -0.4) -- (5.8, -1.1);
\node[font=\tiny, gray] at (6.5, -0.75) {Hamming};

\node[font=\scriptsize, align=center, fill=accentblue!8, draw=accentblue!40, rounded corners=2pt, inner sep=3pt]
    at (4.5, -1.7) {$\mathcal{C}_L$: \texttt{011011010...}};
\node[font=\large] at (5.8, -1.7) {$\oplus$};
\node[font=\scriptsize, align=center, fill=accentgreen!8, draw=accentgreen!40, rounded corners=2pt, inner sep=3pt]
    at (7.1, -1.7) {$\mathcal{C}_R$: \texttt{010011110...}};

\draw[thick, gray!60, -{Stealth[length=3pt]}] (5.8, -2.2) -- (5.8, -2.7);
\node[font=\scriptsize\bfseries, accentorange!80!black, fill=accentorange!8, draw=accentorange!50,
    rounded corners=2pt, inner sep=3pt] at (5.8, -3.1) {popcount(XOR) $=$ cost};

\end{tikzpicture}
\caption{Census Transform encoding and matching. Each pixel's intensity is compared to the center pixel, producing a binary descriptor invariant to monotonic intensity changes. Matching cost is the Hamming distance (popcount of XOR) between left and right descriptors. Green cells: intensity $>$ center ($\to 1$); red cells: $\leq$ center ($\to 0$); orange: center pixel.}
\label{fig:census_transform}
\end{figure}
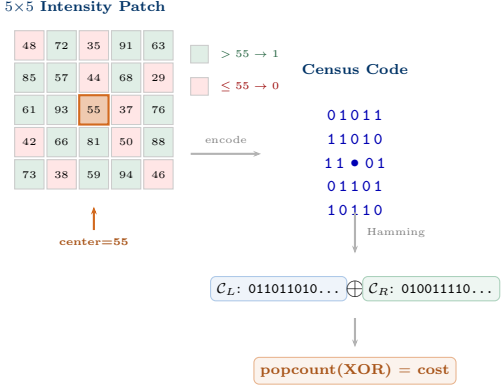

The Census Transform is computed entirely on GPU using a custom CUDA kernel.
Each thread block processes a tile of the image using shared memory for the local intensity patch, avoiding redundant global memory accesses.
Figure~\ref{fig:census_transform} illustrates the encoding and matching process.

\begin{algorithm}[t]
\caption{GPU Census Transform Kernel}
\label{alg:census}
\begin{algorithmic}[1]
\Require Source image $I$, block size $B$
\State \textbf{Grid}: $\lceil W/B \rceil \times \lceil H/B \rceil$ blocks
\State \textbf{Shared}: $(B + 2S_x) \times (B + 2S_y)$ patch
\State Load patch from global memory with halo
\State \texttt{\_\_syncthreads()}
\For{each pixel $(x,y)$ in tile}
    \State $c \gets 1$ \Comment{sentinel bit}
    \For{$i = -S_x$ \textbf{to} $S_x$, $j = -S_y$ \textbf{to} $S_y$}
        \State $v \gets [I(x+i, y+j) > I(x,y)]$
        \State $c \gets (c \ll 1) \mid v$
    \EndFor
    \State $\mathcal{C}(x,y) \gets c$
\EndFor
\end{algorithmic}
\end{algorithm}

The kernel supports multi-resolution: when the Census image dimensions differ from the source, coordinates are scaled accordingly, enabling efficient processing at reduced resolution without a separate resize step.

\subsubsection{Matching Cost and Sub-Pixel Refinement}

Given Census descriptors $\mathcal{C}_L$ and $\mathcal{C}_R$ for left and right images, the matching cost for a set of query points $\mathcal{P}$ at offset $(dx, dy)$ is the sum of Hamming distances:
\begin{multline}
\label{eq:census_cost}
S(\mathcal{P}, dx, dy) = \\
\sum_{(x,y) \in \mathcal{P}} \text{popcount}\big(\mathcal{C}_L(x,y) \oplus \mathcal{C}_R(x{+}dx, y{+}dy)\big)
\end{multline}

The optimal integer offset $dx^*$ is found by exhaustive search within the configured disparity range $[dx_{\min}, dx_{\max}]$ and vertical range $[dy_{\min}, dy_{\max}]$.
The search is parallelized on GPU: each CUDA block handles one query group, with threads cooperatively evaluating different $(dx, dy)$ candidates, followed by a parallel reduction to find the minimum cost.

Sub-pixel refinement is achieved through parabolic interpolation on the neighboring costs:
\begin{equation}
\label{eq:subpixel}
\hat{d}_x = dx^* - \frac{S(dx^*+1) - S(dx^*-1)}{2\big(S(dx^*+1) + S(dx^*-1) - 2S(dx^*)\big)}
\end{equation}
This is computed directly in the CUDA kernel after the parallel reduction, yielding floating-point disparity at negligible additional cost.

\subsubsection{Forward-Backward Verification}

To reject unreliable matches, we employ a \textbf{left-right consistency check} via forward-backward verification.
After finding the best match from left-to-right (forward pass), we perform a right-to-left search (backward pass) from the matched position:

\begin{enumerate}[leftmargin=*,nosep]
\item \textbf{Forward}: Query Census values from the left image at sampled points. Search the right image for the best match, obtaining offset $(dx_f, dy_f)$.
\item \textbf{Backward}: From the matched positions in the right image, query Census values. Search the left image for the best match, obtaining verification offset $(dx_v, dy_v)$.
\item \textbf{Consistency}: Accept the match only if $|dx_v| < \tau_v$, where $\tau_v$ is a pixel threshold. A consistent match should return close to the original position ($dx_v \approx 0$).
\end{enumerate}

This bidirectional verification effectively filters out matches in occluded regions, repetitive textures, and areas with insufficient structure.

\subsubsection{Object-Centric Far/Close Strategy}
\label{sec:far_close}

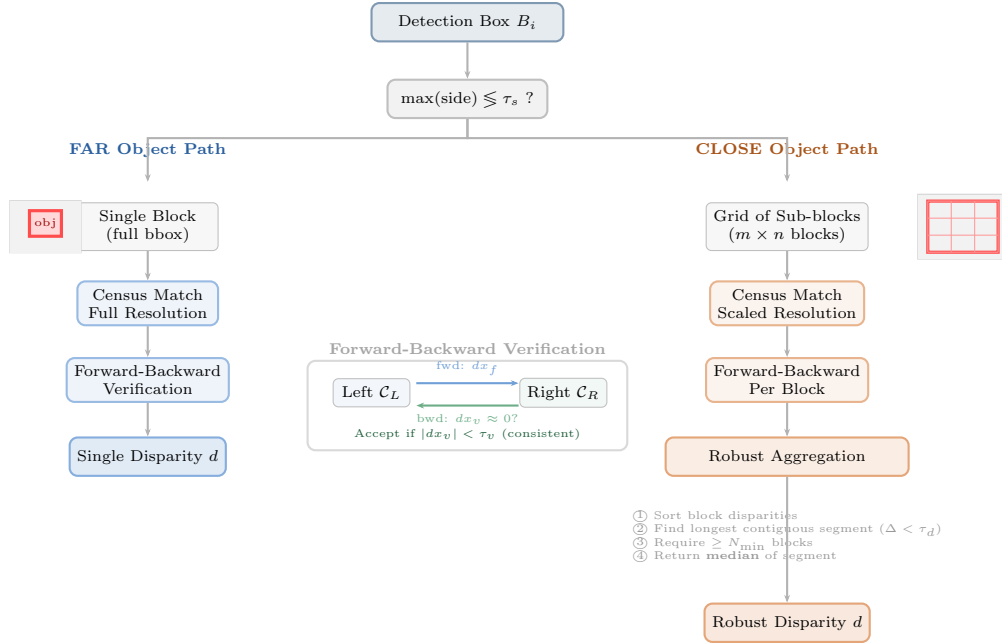
\begin{figure*}[t]
\centering
\begin{tikzpicture}[
    scale=0.85, every node/.style={scale=0.85},
    stgbox/.style={rectangle, rounded corners=3pt, draw, thick, minimum height=0.6cm, font=\scriptsize, align=center},
    datbox/.style={rectangle, rounded corners=2pt, draw=gray!50, fill=gray!6, minimum height=0.5cm, font=\scriptsize, align=center},
    arr/.style={-{Stealth[length=3pt]}, thick, gray!60},
    hlarr/.style={-{Stealth[length=3pt]}, very thick, accentorange!70},
]

\node[stgbox, fill=darkblue!10, draw=darkblue!50, minimum width=3cm] (det) at (0, 0)
    {Detection Box $B_i$};
\node[stgbox, fill=gray!10, draw=gray!50, minimum width=2.5cm] (classify) at (0, -1.2)
    {$\max(\text{side}) \lessgtr \tau_s$ ?};
\draw[arr] (det) -- (classify);

\node[font=\scriptsize\bfseries, accentblue!80!black] at (-5.0, -2.0) {FAR Object Path};
\draw[arr] (classify.south) -- ++(-0,-0.3) -| (-5.0, -2.5);

\node[datbox, minimum width=2.2cm] (far_box) at (-5.0, -3.2) {Single Block\\(full bbox)};
\begin{scope}[shift={(-7.0, -3.2)}]
    \draw[gray!30, thick] (-0.15, -0.4) rectangle (0.95, 0.4);
    \fill[gray!10] (-0.15, -0.4) rectangle (0.95, 0.4);
    \draw[red!70, very thick, fill=red!15] (0.15, -0.15) rectangle (0.65, 0.25);
    \node[font=\tiny, red!60!black] at (0.4, 0.05) {obj};
\end{scope}

\node[stgbox, fill=accentblue!8, draw=accentblue!50, minimum width=2.2cm] (far_census) at (-5.0, -4.4)
    {Census Match\\Full Resolution};
\draw[arr] (far_box) -- (far_census);

\node[stgbox, fill=accentblue!8, draw=accentblue!50, minimum width=2.2cm] (far_fb) at (-5.0, -5.6)
    {Forward-Backward\\Verification};
\draw[arr] (far_census) -- (far_fb);

\node[stgbox, fill=accentblue!15, draw=accentblue!60, minimum width=2.2cm] (far_out) at (-5.0, -6.8)
    {Single Disparity $d$};
\draw[arr] (far_fb) -- (far_out);

\node[font=\scriptsize\bfseries, accentorange!80!black] at (5.0, -2.0) {CLOSE Object Path};
\draw[arr] (classify.south) -- ++(-0,-0.3) -| (5.0, -2.5);

\node[datbox, minimum width=2.4cm] (close_box) at (5.0, -3.2) {Grid of Sub-blocks\\($m \times n$ blocks)};
\begin{scope}[shift={(7.2, -3.2)}]
    \draw[gray!30, thick] (-0.15, -0.5) rectangle (1.25, 0.5);
    \fill[gray!10] (-0.15, -0.5) rectangle (1.25, 0.5);
    \draw[red!70, very thick] (0.0, -0.4) rectangle (1.1, 0.4);
    \foreach \x in {0.0, 0.367, 0.733, 1.1} {
        \draw[red!40, thin] (\x, -0.4) -- (\x, 0.4);
    }
    \foreach \y in {-0.4, -0.133, 0.133, 0.4} {
        \draw[red!40, thin] (0.0, \y) -- (1.1, \y);
    }
\end{scope}

\node[stgbox, fill=accentorange!8, draw=accentorange!50, minimum width=2.4cm] (close_census) at (5.0, -4.4)
    {Census Match\\Scaled Resolution};
\draw[arr] (close_box) -- (close_census);

\node[stgbox, fill=accentorange!8, draw=accentorange!50, minimum width=2.4cm] (close_fb) at (5.0, -5.6)
    {Forward-Backward\\Per Block};
\draw[arr] (close_census) -- (close_fb);

\node[stgbox, fill=accentorange!12, draw=accentorange!60, minimum width=3.8cm] (close_agg) at (5.0, -6.8)
    {Robust Aggregation};
\draw[arr] (close_fb) -- (close_agg);

\begin{scope}[shift={(5.0, -8.2)}]
    \node[font=\tiny, align=left, text=gray!80] at (0, 0.15) {
        \textcircled{\tiny 1} Sort block disparities\\
        \textcircled{\tiny 2} Find longest contiguous segment ($\Delta < \tau_d$)\\
        \textcircled{\tiny 3} Require $\geq N_{\min}$ blocks\\
        \textcircled{\tiny 4} Return \textbf{median} of segment};
\end{scope}

\node[stgbox, fill=accentorange!15, draw=accentorange!60, minimum width=2.4cm] (close_out) at (5.0, -9.4)
    {Robust Disparity $d$};
\draw[arr] (close_agg.south) -- ++(0, -0.3) -- (close_out.north);

\begin{scope}[shift={(0, -5.6)}]
    \node[font=\scriptsize\bfseries, gray!70] at (0, 0.5) {Forward-Backward Verification};
    \draw[gray!30, thick, rounded corners=3pt] (-2.5, -1.1) rectangle (2.5, 0.3);
    \node[datbox, fill=accentblue!6, minimum width=1.2cm, minimum height=0.4cm] (li) at (-1.5, -0.2) {Left $\mathcal{C}_L$};
    \node[datbox, fill=accentgreen!6, minimum width=1.2cm, minimum height=0.4cm] (ri) at (1.5, -0.2) {Right $\mathcal{C}_R$};
    \draw[accentblue!70, thick, -{Stealth[length=3pt]}] (-0.8, -0.05) -- node[above, font=\tiny] {fwd: $dx_f$} (0.8, -0.05);
    \draw[accentgreen!70, thick, -{Stealth[length=3pt]}] (0.8, -0.4) -- node[below, font=\tiny] {bwd: $dx_v \approx 0$?} (-0.8, -0.4);
    \node[font=\tiny, accentgreen!70!black] at (0, -0.85) {Accept if $|dx_v| < \tau_v$ (consistent)};
\end{scope}

\end{tikzpicture}
\caption{Template Matching far/close dual-track strategy and forward-backward verification. Far objects (small bounding boxes) are matched as a single block at full resolution. Close objects (large bounding boxes) are subdivided into a grid, matched at scaled resolution, and robustly aggregated via sorted contiguous segment median. Both paths employ forward-backward verification (center inset): a match is accepted only if the backward search returns to near the original position.}
\label{fig:far_close}
\end{figure*}

Figure~\ref{fig:far_close} illustrates the dual-track processing strategy for far and close objects.
A key design decision is the separation of objects into \textbf{far} and \textbf{close} categories based on their apparent size in the image:

\begin{equation}
\text{type} = \begin{cases}
\text{FAR} & \text{if } \max(w \cdot W, h \cdot H) < \tau_s \\
\text{CLOSE} & \text{otherwise}
\end{cases}
\end{equation}
where $(w, h)$ is the normalized bounding box size, $(W, H)$ is the image resolution, and $\tau_s$ is the minimum side length threshold.

\paragraph{Far Objects.}
For distant objects (small bounding boxes), the entire box is treated as a single matching block.
Query points are sampled on a regular grid within the box (with configurable side point count and maximum total points).
Matching is performed on the \emph{original resolution} Census image to preserve the fine detail needed for small disparities.
A single disparity value per object is obtained directly from the block match.

\paragraph{Close Objects.}
For nearby objects (large bounding boxes), the box is subdivided into a grid of smaller blocks, each matched independently.
Matching is performed on a \emph{downscaled} Census image (controlled by \texttt{image\_scale\_for\_close\_objects}) with proportionally wider search ranges, since close objects have large disparities that can be accurately captured at lower resolution.

The per-block disparities are then robustly aggregated:
\begin{enumerate}[leftmargin=*,nosep]
\item Sort all valid block disparities
\item Find the longest contiguous segment where consecutive differences are below threshold $\tau_d$
\item Require the segment to contain at least $N_{\min}$ blocks
\item Return the \emph{median} of the segment as the object disparity
\end{enumerate}

This aggregation rejects outlier blocks (from background visible through windows, occluding objects, or reflection artifacts) while capturing the dominant in-object disparity.

\begin{algorithm}[t]
\caption{Template Match: Object Disparity Estimation}
\label{alg:template_match}
\begin{algorithmic}[1]
\Require Left/right GPU images, detection boxes $\{B_i\}$
\State Compute Census images at original and scaled resolutions
\For{each detection $B_i$}
    \If{$\text{maxSide}(B_i) < \tau_s$}
        \State Classify as FAR; create single query block
    \Else
        \State Classify as CLOSE; subdivide into grid of blocks
    \EndIf
    \State Sample query points, excluding occluded regions
\EndFor
\State \textbf{Batch match} all FAR blocks on original Census images
\State \textbf{Batch match} all CLOSE blocks on scaled Census images
\For{each query block}
    \State Forward match $\to$ $(dx, dy, \text{score})$
    \State Backward verify $\to$ $(dx_v, dy_v)$
    \If{$|dx_v| < \tau_v$}
        \State Record sub-pixel disparity via Eq.~\ref{eq:subpixel}
    \EndIf
\EndFor
\For{each FAR object}
    \State Disparity $\gets$ single block result
\EndFor
\For{each CLOSE object}
    \State Disparity $\gets$ robust aggregation of block disparities
\EndFor
\end{algorithmic}
\end{algorithm}

\subsubsection{Occlusion-Aware Point Sampling}

Before matching, the system identifies \emph{occluding} objects for each detection: an object $B_j$ occludes $B_i$ if $B_j$ overlaps $B_i$ in the image and $B_j$ has a larger bottom $y$-coordinate (closer to the camera in perspective projection).
Query points that fall within any occluding object's bounding box are excluded from sampling.
If too many candidate objects exist (exceeding a configurable maximum), a priority-based selection favoring frontal objects (within a central crop region) is applied, with the remaining slots filled by proximity-ordered objects.

\subsubsection{Disparity-to-Depth Conversion}

Given the matched disparity $d$ at image coordinates $(u, v)$ and stereo projection matrices $P_1, P_2$, the 3D point in camera frame is recovered using the $Q$ matrix (from stereo rectification) or direct triangulation:
\begin{equation}
\begin{bmatrix} X \\ Y \\ Z \\ W \end{bmatrix} = Q \begin{bmatrix} u \\ v \\ d \\ 1 \end{bmatrix}, \quad \mathbf{p}_{\text{cam}} = \frac{1}{W}\begin{bmatrix} X \\ Y \\ Z \end{bmatrix}
\end{equation}
The point is then transformed to the IMU/vehicle frame via the extrinsic calibration $T_{\text{cam} \to \text{imu}}$.

\subsubsection{Covariance Estimation}

The uncertainty of the stereo depth estimate is propagated from disparity uncertainty.
Given disparity variance $\sigma_d^2$, the depth variance is:
\begin{equation}
\sigma_Z^2 = \left(\frac{\partial Z}{\partial d}\right)^2 \sigma_d^2 = \frac{Z^4}{f^2 b^2} \sigma_d^2
\end{equation}

For the Template Match method, the disparity variance $\sigma_d^2$ is set to a configured value (\texttt{tm\_stereo\_disparity\_variance}).
For dense BM/SGM methods, a dynamic variance model is employed:
\begin{equation}
\label{eq:dynamic_variance}
\sigma_d^2 = \frac{\sigma_{\text{obs}}^2}{n_{\text{near}}} + \gamma (\bar{d}_{\text{near}} - \bar{d}_{\text{all}})^2 + \sigma_{\text{sys}}^2
\end{equation}
where $n_{\text{near}}$ is the number of nearby-percentile disparity samples, $\bar{d}_{\text{near}}$ and $\bar{d}_{\text{all}}$ are the mean disparities of near samples and all samples respectively, $\sigma_{\text{obs}}^2$ is a configured observation variance, $\gamma$ is a scaling factor, and $\sigma_{\text{sys}}^2$ is the system noise floor.
This formulation captures both measurement noise (first term) and the spread of disparities within the object (second term, indicating potential depth ambiguity from mixed foreground/background samples).

The full 3D covariance is obtained by rotating the camera-frame covariance through extrinsic rotation $R$:
\begin{equation}
\Sigma_{\text{imu}} = R \cdot \text{diag}(\sigma_x^2, \sigma_y^2, \sigma_Z^2) \cdot R^\top
\end{equation}

\section{Online Calibration Refinement}
\label{sec:calibration}

Stereo systems in production vehicles are subject to continuous mechanical vibration, thermal expansion, and settling, causing the extrinsic calibration to drift over time.
Our system addresses this through three complementary online refinement mechanisms.

\subsection{Auto Rectification Offset}
\label{sec:auto_rect}

Vertical misalignment between stereo cameras directly corrupts disparity estimation.
The auto-rectification module searches over integer vertical offsets $\delta \in [\delta_{\min}, \delta_{\max}]$ to find the alignment that maximizes stereo matching quality:

\begin{equation}
\delta^* = \arg\max_{\delta} \; \#\{(x,y) : d(x, y; \delta) > d_{\min}\}
\end{equation}

where $d(x,y;\delta)$ is the BM disparity at $(x,y)$ when the left image is shifted vertically by $\delta$ pixels.
The search is performed over a designated ROI and runs asynchronously on separate CUDA streams---one per candidate offset---enabling parallel evaluation.

The raw per-frame estimates are temporally filtered using a \textbf{sliding window median filter} to prevent sudden jumps:
\begin{gather}
\delta_t = \text{median}\{\delta^*_{t-k}, \ldots, \delta^*_t\} \\
\Delta\delta_t = \text{clamp}(\delta_t - \delta_{t-1},\; \pm\Delta_{\max})
\end{gather}

\subsection{Radar Disparity Refiner}
\label{sec:radar_refiner}

Radar provides absolute range measurements that serve as reference for correcting systematic stereo disparity bias.
The refiner projects radar detections into disparity space using the composite projection:
\begin{equation}
\mathbf{p}_{\text{disp}} = Q^{-1} T_{\text{imu} \to \text{cam}} \mathbf{p}_{\text{radar}}
\end{equation}

For each radar detection, the system compares the predicted radar disparity with the actual stereo disparity within a bounding box derived from the detection's 3D extent projected to image coordinates.
The difference is accumulated into a voting histogram spanning $\pm K$ pixels at sub-pixel resolution (16 sub-levels per pixel):

\begin{algorithm}[t]
\caption{Radar Disparity Refiner}
\label{alg:radar_refiner}
\begin{algorithmic}[1]
\Require Disparity map $D$, radar detections $\{r_k\}$
\State $\mathbf{v} \gets \mathbf{0}$ \Comment{Vote vector, dimension $2K \cdot 16 + 1$}
\For{each radar detection $r_k$}
    \State Project $r_k$ to image; get bounding box and $d_{\text{radar}}$
    \State Find offset closest to zero within box: $\Delta d_k$
    \State $\mathbf{v}[\Delta d_k] \mathrel{+}= 1$ \Comment{Saturate at 1 per detection}
\EndFor
\State $\mathbf{v} \gets \text{GaussianSmooth}(\mathbf{v}, \sigma=1)$
\State $\mathbf{m}_t \gets (1-\lambda)\mathbf{m}_{t-1} + \lambda \hat{\mathbf{v}}$ \Comment{EMA update}
\State $\Delta d^* \gets \arg\max \mathbf{m}_t$
\State $\bar{\Delta d}_t \gets (1-\lambda)\bar{\Delta d}_{t-1} + \lambda \Delta d^*$ \Comment{Smooth offset}
\State Apply $\text{clamp}(\bar{\Delta d}_t, \pm 3\text{px})$ to disparity map
\end{algorithmic}
\end{algorithm}

The voting scheme uses saturation (at most 1 vote per detection) to prevent large foreground objects from dominating.
A Gaussian kernel ($\sigma = 1$) smooths the vote histogram as a simple approximation to kernel density estimation.
The vote memory is updated via exponential moving average (EMA) with rate $\lambda$, providing temporal stability while adapting to drift.

\subsection{Object Disparity Refiner}
\label{sec:obj_refiner}

For the Template Match pathway, the Object Disparity Refiner provides a complementary calibration signal at the object level.
It associates stereo-derived 3D positions with radar detections through a two-stage process:

\paragraph{Stage 1: Coarse Search.}
For each candidate disparity offset, all stereo object points are projected to the vehicle frame and associated with the nearest radar point (found via KD-tree search within a configurable range).
A greedy 1-to-1 bipartite matching is used: point pairs are sorted by distance, and each stereo/radar point participates in at most one pair.
The score for each offset combines association quality with a temporal consistency bonus:
\begin{multline}
\text{score}(\Delta d) = \sum_{(s,r) \in \mathcal{M}} \max\!\left(1 - \frac{\|s - r\|}{R_{\max}}, 0\right) \\
+ \beta \cdot \mathbb{1}[\Delta d = \Delta d_{t-1}]
\end{multline}

\paragraph{Stage 2: Iterative Refinement.}
Starting from the coarse estimate, the offset is refined by averaging disparity differences from well-matched pairs (within a tolerance band), with a prior term pulling toward the previous frame's offset:
\begin{equation}
\Delta d_{\text{ref}} = \frac{\sum_{i} \Delta d_i \cdot \mathbb{1}_{|\Delta d_i - \hat{\Delta d}| < \tau} + w_p \Delta d_{t-1}}{\sum_{i} \mathbb{1}_{|\Delta d_i - \hat{\Delta d}| < \tau} + w_p}
\end{equation}
where $\hat{\Delta d}$ is the coarse estimate and $w_p$ is the prior weight.

The final offset is rate-limited between consecutive frames to prevent abrupt changes.

\section{Vision Tracking and Fusion}
\label{sec:tracking}

\subsection{Detection-to-Track Association}

The vision tracker maintains a set of active tracklets, each representing a detected vehicle across time.
Association between new detections and existing tracks uses the \textbf{Hungarian algorithm} on a combined cost matrix:
\begin{equation}
\text{cost}(d_i, t_j) = \text{IoU}(B_i, B_j) + \frac{\mathbf{f}_i \cdot \mathbf{f}_j}{\|\mathbf{f}_i\| \|\mathbf{f}_j\|}
\end{equation}
where $B_i, B_j$ are bounding boxes, and $\mathbf{f}_i, \mathbf{f}_j$ are CNN feature vectors extracted at the detection centers via bilinear interpolation from the detection network's feature map.

Association uses gating: pairs with zero IoU, large center displacement, or excessive longitudinal separation in 3D are rejected.

\subsection{State Estimation}

Each \texttt{DefaultVisionTracklet} maintains a Kalman filter with state vector:
\begin{equation}
\mathbf{x} = [x_{\text{rel}}, \; y_{\text{rel}}, \; v_x, \; v_y]^\top
\end{equation}
where $(x_{\text{rel}}, y_{\text{rel}})$ is the longitudinal and lateral distance to the ego vehicle, and $(v_x, v_y)$ are relative velocities.
The prediction step accounts for ego motion through odometry integration.

The system supports two measurement models:
\begin{itemize}[leftmargin=*,nosep]
\item \textbf{With speed}: Full state update when vision speed estimation is available.
\item \textbf{Without speed}: Position-only update as fallback.
\end{itemize}

\subsection{Vehicle Geometry Estimation}

For large vehicles (trucks, buses), the tracklet maintains estimates of vehicle geometric properties (e.g., trailer width) using \textbf{Iteratively Reweighted Least Squares (IRLS)}.
This robust estimator down-weights outlier measurements, providing stable geometry estimates over time that improve ranging accuracy for partially visible vehicles.

\subsection{Vision Speed Estimation}

Vehicle speed is estimated by observing the change in stereo-derived longitudinal distance over time, compensated for ego motion:
\begin{equation}
v_{\text{obj}} = v_{\text{ego}} + \frac{\Delta x_{\text{rel}}}{\Delta t}
\end{equation}

The speed estimate is cross-validated between stereo-based and monocular methods, with a reliability status that gates its contribution to downstream modules (e.g., congestion detection).

\subsection{Multi-Source Depth Fusion}

The final depth for each detection is selected from multiple candidates with a priority scheme:
\begin{enumerate}[leftmargin=*,nosep]
\item Refined stereo (with face-aware keypoint refinement)
\item Stereo disparity (BM/SGM or Template Match)
\item Ground point triangulation
\item Apparent size estimation
\end{enumerate}

A \textbf{sanity check} cross-validates stereo results against monocular estimates: if the stereo depth differs from the monocular depth by more than a configured ratio (accounting for the expected object size), the stereo result is rejected.
This prevents catastrophic errors from mismatches while preserving stereo's superior accuracy when it agrees with geometric priors.

\section{System Design Discussion}
\label{sec:discussion}

\subsection{GPU Pipeline and Latency}

The system exploits CUDA streams extensively to overlap computation.
The typical per-frame timeline is:
\begin{enumerate}[leftmargin=*,nosep]
\item \texttt{preprocessAsync}: Color-to-gray conversion, downsampling (non-blocking)
\item \texttt{computeAsync}: Launch BM/SGM or Template Match (non-blocking)
\item \texttt{computeRectificationOffsetAsync}: Parallel offset search (non-blocking)
\item \texttt{DetectAsync}: CNN inference (non-blocking, separate stream)
\item \texttt{waitForCompletion}: Synchronize all streams
\item Sequential: Closest point computation, track, fusion
\end{enumerate}

Template Matching is particularly efficient because it avoids computing disparity for the entire image.
For a typical highway frame with 20--50 detected objects, Template Matching processes $O(10^4)$ query points compared to $O(10^6)$ pixels for dense methods.

\subsection{Comparison of Stereo Methods}

Table~\ref{tab:comparison} summarizes the characteristics of the three stereo methods.

\begin{table}[t]
\centering
\caption{Comparison of stereo matching methods.}
\label{tab:comparison}
\footnotesize
\begin{tabular}{@{}lccc@{}}
\toprule
\textbf{Property} & \textbf{BM} & \textbf{SGM} & \textbf{Template} \\
\midrule
Output & Dense map & Dense map & Per-object \\
Matching cost & SAD & Census/MI & Census \\
Regularization & None & 4-dir path & F-B verify \\
Radiometric inv. & Low & Medium & High \\
Long-range acc. & Low & Medium & High \\
Computation & High & High & Low \\
Calibration ref. & Rect.+Radar & Rect.+Radar & Obj.Refiner \\
\bottomrule
\end{tabular}
\end{table}

\subsection{Template Match: Key Innovations}

Summarizing the novel aspects of the Census-based Template Matching approach:

\begin{enumerate}[leftmargin=*,nosep]
\item \textbf{Object-centric sparse matching}: Computes disparity only where needed, reducing computation by 1--2 orders of magnitude compared to dense methods.
\item \textbf{Census Transform on GPU}: Custom CUDA kernel with shared memory optimization and multi-resolution support, providing robustness to radiometric differences (gain, exposure, white balance) between left and right cameras.
\item \textbf{Far/close divide-and-conquer}: Adapts resolution and matching strategy to object distance---full resolution for far objects where sub-pixel accuracy is critical, downscaled multi-block for close objects where disparity is large.
\item \textbf{Forward-backward verification}: Bidirectional consistency check filters unreliable matches without requiring a full disparity map.
\item \textbf{Occlusion-aware sampling}: Excludes query points in regions occluded by closer objects, preventing background contamination.
\item \textbf{Robust multi-block aggregation}: For close objects, the sorted-contiguous-segment algorithm rejects outlier blocks from windows, reflections, and background, using the median of the dominant cluster for stability.
\item \textbf{Priority-based object selection}: When detection count exceeds the budget, frontal objects are prioritized to ensure ranging accuracy for the most safety-critical targets.
\item \textbf{Radar-stereo object-level refinement}: The Object Disparity Refiner provides continuous calibration correction specific to the Template Match pathway.
\end{enumerate}

\subsection{Stereo Vision as Pseudo-LiDAR: Broader Insights}

The development trajectory of our system validates the ``pseudo-LiDAR'' thesis~\cite{wang2019pseudo}---that stereo cameras can functionally replace LiDAR for many autonomous driving tasks when paired with sufficiently sophisticated algorithms.
Several deeper insights emerge from production deployment:

\paragraph{The ``Detect First vs. Measure First'' Dichotomy.}
Monocular depth estimation follows a ``detect first, then infer depth'' paradigm: the system must recognize an object's class before it can estimate distance (via learned size priors or depth networks).
This creates a fundamental vulnerability to non-standard obstacles---fallen cargo, construction debris, or novel vehicle types that were absent from training data.
Stereo vision inverts this paradigm: \emph{depth measurement requires no recognition}.
The disparity of a pixel depends only on its correspondence between left and right views, not on what object it belongs to.
This ``measure first'' capability is critical for safety, as it enables ranging of arbitrary obstacles that no learned model has seen before.

\paragraph{Why Long-Range Stereo Requires Object-Centric Design.}
The classical dense stereo paradigm fails gracefully at close range but catastrophically at long range.
At 200m, a vehicle subtends perhaps $50 \times 30$ pixels, and its true disparity is 2--3 pixels.
A dense disparity map at this scale is dominated by noise, texture artifacts, and sky/road confusion.
Our Template Matching approach succeeds because it \emph{aggregates} evidence from all pixels within the detection box.
Even if individual pixel matches are noisy, the collective vote of 100--500 Census-matched points within the box converges to a reliable disparity estimate.
This is analogous to the signal-processing principle that averaging $N$ independent measurements reduces noise by $\sqrt{N}$.

\paragraph{Complementarity with Other Sensors.}
In practice, no single sensor is sufficient for all conditions.
Our system design reflects this reality: radar provides absolute range references for calibration refinement (Section~\ref{sec:radar_refiner}), monocular priors serve as sanity checks (Section~\ref{sec:tracking}), and dense disparity maps capture scene-level structure when computational budget allows.
The Template Match module serves as the \emph{primary} long-range ranging sensor, while radar and mono provide complementary supervision signals.
This multi-modal co-design---where each sensor's strength compensates for another's weakness---is the pragmatic path toward robust autonomous perception.

\paragraph{From Dense Stereo to BEV Fusion.}
Industry trends are converging toward BEV (Bird's Eye View) representations that unify multi-camera, LiDAR, and radar data in a common spatial frame.
Stereo depth---whether from dense SGM or sparse Template Matching---naturally feeds into BEV feature construction by providing per-pixel or per-object depth distributions.
Our object-level disparities can be directly projected into BEV space as high-confidence depth anchors, guiding the lifting of 2D features to 3D.
This positions the Template Match approach as a bridge between classical geometric stereo and modern learned BEV architectures.

\section{Conclusion}
\label{sec:conclusion}

We have presented a comprehensive stereo ranging system for autonomous driving that integrates multiple complementary depth estimation methods within a unified detection-ranging-calibration-tracking pipeline.
At its core, the system addresses the fundamental question of how stereo vision---the ``pseudo-LiDAR'' of the perception stack---can best serve the needs of autonomous driving, particularly for the critical long-range regime where LiDAR point density is sparse and monocular depth estimation becomes unreliable.

The Census-based Template Matching algorithm represents a paradigm shift from dense disparity computation to object-centric sparse matching, achieving superior accuracy at long range with significantly reduced computational cost.
The combination of Census Transform's radiometric invariance, GPU-accelerated parallel matching, forward-backward verification, and robust multi-block aggregation addresses the key challenges of production stereo systems.
Crucially, by decoupling depth measurement from object recognition, the system can range arbitrary obstacles---including non-standard objects absent from any training dataset---fulfilling stereo vision's unique role as a physics-grounded, category-agnostic depth sensor.

The multi-layered online calibration refinement framework---combining auto-rectification, radar-based voting, and object-level association---ensures sustained accuracy despite the mechanical vibration and thermal drift inherent in vehicle-mounted sensors.
Through asynchronous GPU pipeline design and careful engineering of the detection-to-tracking data flow, the system meets the real-time constraints of highway autonomous driving while maintaining the flexibility to evolve toward emerging BEV-centric fusion architectures.

\bibliographystyle{plain}

\end{document}